\documentclass{article}
\usepackage{spconf}
\usepackage{color}
\usepackage{math}
\usepackage{enumitem}
\usepackage{theorem}
\usepackage{pifont}
\usepackage{multirow}
\usepackage{hyperref}
\usepackage{cite}
\usepackage{graphicx}
\usepackage{amsmath}
\usepackage{amsfonts}
\usepackage{amssymb}
\usepackage{subfig}
\usepackage{caption}
\usepackage{amsopn}
\usepackage{xspace}
\usepackage{array}
\usepackage{epsfig}
\usepackage{algorithm}
\usepackage{algorithmic}
\usepackage{float}

\usepackage{cleveref}

\crefformat{footnote}{#2\footnotemark[#1]#3}

\title{Blind Source Separation Using Mixtures of Alpha-Stable Distributions}

\name{\vspace{-4mm}Nicolas Keriven${}^*$, Antoine Deleforge${}^*$ and Antoine Liutkus${}^\dagger$\thanks{This work was partly supported by the research programme KAMoulox (ANR-15-CE38-0003-01) funded by ANR, the French State agency for research.}}
\address{${}^*$Inria Rennes - Bretagne Atlantique, France\\
         ${}^\dagger$Inria and LIRMM, University of Montpellier, France}
\begin{document}
\ninept
\maketitle
\begin{abstract}
We propose a new blind source separation algorithm based on mixtures of $\alpha$-stable distributions. Complex symmetric $\alpha$-stable distributions have been recently showed to better model audio signals in the time-frequency domain than classical Gaussian distributions thanks to their larger dynamic range. However, inference with these models is notoriously hard to perform because their probability density functions do not have a closed-form expression in general. Here, we introduce a novel method for estimating mixtures of $\alpha$-stable distributions based on characteristic function matching. We apply this to the blind estimation of binary masks in individual frequency bands from multichannel convolutive audio mixtures. We show that the proposed method yields better separation performance than Gaussian-based binary-masking methods. 
\end{abstract}

\section{Introduction}
This paper is concerned with source separation, which is a topic in
applied mathematics that aims at processing~\emph{mixture} signals
so as to recover their constitutive components, called \emph{sources}~\cite{ss_review2010}.
It is a field of important and widespread practical applications,
notably in audio. It is traditionally exemplified by the \emph{cocktail
party problem}, which consists in isolating some specific discussion within
the recording of a crowd~\cite{Yilmaz2004,rickard2007duet}. Apart
from such speech processing scenarios, source separation also enjoyed
much interest in the music processing literature, due to its important
applications in the entertainment industry~\cite{vincentSSoverview2014}. 

From the perspective of this paper, it is worth mentioning that a
significant portion of the research on source separation first makes
some \emph{assumptions} on the source signals and then picks some
\emph{mixing model}. While the former usually stands on probabilistic
grounds, the latter often comes from physical assumptions and explains
how the observed mixtures are generated from the sources.

Historically, the \emph{overdetermined} linear case was considered, \textit{i.e.},
more mixtures than sources are available~\cite{ss_review2010}.
The interesting fact about such mixing models is they can be inverted
easily, allowing to recover the sources from the mixtures, provided
their parameters are known. The breakthrough brought in by source
separation is to allow identification of such mixing parameters with
only very general assumptions about the sources. These assumptions
are mostly either that sources are independent, identically distributed
(iid.) and non-Gaussian, as in Independent Components Analysis (ICA,~\cite{hyvarinen2004independent}),
or that they are Gaussian but not iid. as in Second-Order Blind
Identification (SOBI~\cite{belouchrani1997blind}). Going in the
frequency domain allowed to extend such approaches to \emph{convolutive}
mixtures, i.e. for which the sensors capture the sources after some
acoustic propagation whose duration is not negligible.

The validity of the mixing model and its invertibility is crucial
for applying separation methods that make only broad assumptions on
the sources. When such assumptions are violated, those approaches
are not applicable. This typically happens in the underdetermined
scenario, where fewer mixtures than sources are available, which is
common in audio. In that case, separation may only be achieved
through more involved source models and \emph{time-varying} filtering
procedures~\cite{vincentSSoverview2014}. For this reason, it is
natural that research in underdetermined separation focused on highly
parameterized and tractable source models. In short, a huge part of
the models proposed in the literature stands on Gaussian grounds,
where one wants to estimate time-varying power-spectral densities
and steering vectors for building the corresponding multichannel Wiener
filters~\cite{benaroya-separation06,duong_TSALP2010}.
In that framework, estimation is typically achieved through maximization
of likelihoods, for instance using the Expectation-Maximization (EM)
algorithm~\cite{EM-superimposedFeder}. This line of thought leaves
room for much flexibility and a large community strived to provide
effective audio spectrogram models, from sophisticated linear factorization~\cite{ozerov_generalframework2011}
to recent developments in deep learning~\cite{nugraha2016multichannel}.

An intrinsic weakness of Gaussian processes for modelling audio
sources is to require many parameters to faithfully represent sophisticated
signals. This is made unavoidable by their light-tails, which only
allow for small explorations around averages and standard deviations.
One typically has to pick a different Gaussian distribution for \emph{each}
time-frequency bin to obtain a good model~\cite{duong_TSALP2010},
and precise estimation of \emph{all} parameters is required for good
performance. This inevitably makes all related estimation methods
very sensitive to initialization. Using distributions with heavier-tails
than the Gaussian for underdetermined separation has been less explored \cite{yoshii2016student}
although it is common practice in the overdetermined case~\cite{kidmose2001blind,kidmose2001independent}.
Among such distributions, the~$\alpha$-stable distribution~\cite{samoradnitsky1994stable}
enjoyed some interest in signal processing~\cite{nikias1995signal}
and more particularly in source separation recently, because it was
shown to straightforwardly yield effective filters with better perceived
audio quality than the more classical Wiener~\cite{harmonizable2015,fontaine2017explaining}.

However, the delicate question of how to estimate the parameters of
$\alpha$-stable source models remains quite an open issue. It appears to be
very challenging because such distributions do not provide an
analytical expression for their likelihood, which prevents the use of classical
inference methods. Two alternative options were considered so far. First,
Markov Chain Monte Carlo methods~\cite{csimcsekli2015alpha} are
applicable and effective at yet a high computational cost. Second,
classical moment-matching methods were proposed~\cite{liutkusFIM2015}
that are effective, but somewhat ad-hoc and hard to translate into
the \emph{multichannel} case of several mixtures.


In this paper, we use a variant of the recent algorithm introduced in \cite{sketching2016} for the estimation of mixture models by generalized moment matching (GeMM), to exploit mixtures of \emph{multivariate} $\alpha$-stable distributions in the context of audio source separation. This algorithm, referred to as Compressive Learning-Orthogonal Matching Pursuit with Replacement (CL-OMPR), is a greedy, heuristic method that was initially used in the context of \emph{sketching}~\cite{sketching2016}, to estimate mixture models on large-scale databases using only a collection of generalized moments computed in one pass.
Sketching enjoyed several successful applications
in machine learning~\cite{gribonval2017compressive}, but also in source localization~\cite{fontaine2017scalable}. 
Here, we exploit instead the capacity of CL-OMPR to estimate an $\alpha$-stable mixture model whose  probability density function does not enjoy an analytical expression.

\section{Alpha-Stable Unmixing}

\subsection{Alpha-stable mixture model}
Let us consider a mixture of $K$ sound sources observed through $M$ channels. We denote by  $\{s_k(f,t)\}_{k=1}^{K}$ the emitted source spectrograms and by $\{x_m(f,t)\}_{m=1}^{M}$ the observed channel spectrograms in the complex short-time Fourier domain, where $f\in[1\dots F]$ and $t\in[1\dots T]$ denote the discrete frequency and time indexes. Assuming time-domain convolutive filters from sources to channels which are short compared to the Fourier windows, the mixing model at $(f,t)$ can be written
\begin{equation}
\xvect(f,t) = \sum_{k=1}^K \avect_k(f)s_k(f,t)
\end{equation}
where $\xvect(f,t)\in\mathbb{C}^M$ is the observed vector, $\svect(f,t)\in\mathbb{C}^K$ the source vector and $\avect_k(f)\in\mathbb{C}^M$ source $k$'s steering vector.

Now, we choose an original probabilistic model for the source signals, inspired by recent research on~$\alpha$-harmonizable processes~\cite{harmonizable2015,fontaine2017scalable}. For each $f$, all~$\left\{s_k(f,t)\right\}_{t=1}^T$ are assumed independent and identically distributed (iid.) with respect to (wrt.) a symmetric complex and centered $\alpha$-stable distribution of unit scale parameter and characteristic exponent~$\alpha_{k,f}$, which we write:
\begin{equation}
p(s_k(f,t);\alpha_{k,f}) = \mathcal{S}_c(s_k(f,t);\alpha_{kf}).
\label{eqn:model_skft}\end{equation}
In short, the symmetric centered $\alpha$-stable distribution generalizes the Gaussian isotropic one~\cite{circularSymmetricGallager}, while providing significantly heavier tails as its characteristic exponent~$\alpha_{kf}\in]0,2]$ gets small, $\alpha_{kf}=2$ corresponding to the Gaussian case. Contrary to classical Gaussian audio source models~\cite{benaroya-separation06,duong_TSALP2010} the parameters of the proposed model are time-invariant, drastically reducing its size. This is permitted by the fact that the distribution $\mathcal{S}_c$ enables important dynamics for~$s_k(f,t)$. In other words,~\eqref{eqn:model_skft} corresponds to a model for the~\textit{marginal} distribution of the sources. Such ideas have already been considered in~\cite{fontaine2017scalable}. The particularity of our approach in this regard is to feature a \textit{frequency-dependent} characteristic exponent~$\alpha_{kf}$ for increased expressive power. The choice of a unit scale for the distribution comes with no loss of generality: any frequency-dependent scaling of the sources is incorporated in the steering vectors~$\boldsymbol{a}_k(f)$.

We highlight that the probability density function (pdf) of $s_k(f,t)$ in~\eqref{eqn:model_skft} does not have a closed-form expression except for $\alpha_{kf}=1$ (Cauchy) and $\alpha_{kf}=2$ (Gaussian). However, its \textit{characteristic function}, defined as the Fourier transform of its pdf does. We have~\cite{samoradnitsky1994stable,harmonizable2015}:
\begin{equation}
\forall \omega\in\mathbb{C},\mathbb{E}\{\exp(i\text{Re}\left[\omega^\star s_k(f,t)\right])\} = \exp(-|\omega|^{\alpha_{kf}}).
\end{equation}

At this point, we make one important simplifying assumption: we suppose \textit{only one source} is significantly active at each time-frequency (TF) point. More specifically, let~$z(f,t)$ be the index of the source that has the strongest magnitude~$|s_k(f,t)|$ at TF bin~$(f,t)$. Our assumption is that all other sources have a magnitude \textit{close to~$0$}. This is less strong than the so called W-disjoint orthogonality assumption \cite{rickard2007duet} where a single source is assumed to be active. This allows us to assume weak sources are approximately distributed wrt a Gaussian distribution. Indeed, even if it lacks an analytical expression, the pdf for a symmetric $\alpha$-stable distribution is infinitely differentiable close to the origin~\cite{samoradnitsky1994stable}, justifying this second order approximation for weak sources.

As a result of these assumptions, we take our mixture as:
\begin{equation}
\xvect(f,t) = \sum_{k=1}^K \mathbb{I}(z(f,t)=k)\{\avect_k(f)s_k(f,t) + \evect_k(f,t)\},
\end{equation}
where $\mathbb{I}$ is the indicator function and $\evect_k(f,t)\in\mathbb{C}^M$ is a residual Gaussian term containing all non-dominating signals (other than $k$) and possible additional noise.
For convenience, we neglect the interchannel correlations coming from weak sources, to simply assume that~$e_k$  is composed of iid. entries with variance~$\sigma^2_{kf}$:
\begin{equation}
\label{eq:residual}
p(\evect_k(f,t)|z(f,t)=k;\sigma_{kf}^2) = \mathcal{N}_c(\evect_k(f,t);\zerovect,\sigma_{kf}^2\Imat_M)
\end{equation}
where $\mathcal{N}_c$ denotes the multivariate complex circular-symmetric Gaussian distribution~\cite{circularSymmetricGallager}, $\Imat_M$ is the $M-$dimensional identity matrix and $\sigma_{kf}^2$ is the residual variance at frequency $f$ when source $k$ dominates. Furthermore, the indexes $z(f,t)$ of the strongest source for each TF bin are modelled as iid. multinomial variables:
\begin{equation}
\label{eq:assignment}
p(z(f,t)=k;\pivect_f) = \pi_{kf}
\end{equation}
where $\pi_{kf}$ is the probability of source $k$ dominating in frequency band $f$, and $\sum_k\pi_{kf} = 1$.

From all the preceding assumptions and dropping the indexes~$(f,t)$ for convenience, we deduce the characteristic functions of $\avect_ks_k$, $\evect_k$ and $\xvect|z=k$, where~$\omegavect\in\mathbb{C}^M$:
\begin{align}
&\psi_{\footnotesize\avect_ks_k}(\omegavect) = \exp(-|\avect_k^{\star}\omegavect|^{\alpha_{k}}) \\
&\psi_{\footnotesize\evect_k}(\omegavect) = \exp(-\sigma_k^2\|\omegavect\|_2^2) \\
\label{eq:xgivenw}
&\psi_{\footnotesize\xvect|z=k}(\omegavect) = \exp(-|\avect_k^{\star}\omegavect|^{\alpha_{k}}-\sigma_k^2\|\omegavect\|_2^2).
\end{align}
Combining (\ref{eq:assignment}) and (\ref{eq:xgivenw}), we deduce that $\left\{\xvect(f,t)\right\}_{t}$ follows a mixture model parametrized by
\begin{equation}
\label{eq:modelparam}
\thetavect_f = \{\alpha_{kf},\sigma_{kf}^2,\avect_k(f),\pi_{kf}\}^K_{k=1}.
\end{equation}
Following the two-stage approach of \cite{sawada2007two}, the proposed blind source separation method consists in clustering observations $\xvect(f,t)$ independently at each frequency according to this mixture model. The resulting classical source permutation ambiguity across frequencies is left aside here (see Section~\ref{sec:permut}), and a binary mask is then obtained for each source \cite{Yilmaz2004,rickard2007duet}. The special Gaussian case $\alpha_{fk}=2$ is discussed in Section~\ref{sec:alpha2} while a parameter estimation method for the general case is given in Section \ref{sec:sketch}.

\subsection{Special case $\alpha_{fk} = 2$}
\label{sec:alpha2}

Let us consider the special Gaussian case where $\alpha_{fk}=2$ for all $f,k$. The observation model at each frequency becomes
\begin{equation}
p(\xvect_t|z_t=k;\thetavect) = \mathcal{N}_c(\xvect_t;\zerovect,\avect_k\avect_k^\star+\sigma_{k}^2\Imat_M)
\end{equation}
where frequency indexes have been dropped for convenience. The parameters $\thetavect$ of this mixture model can be straightforwardly estimated via an expectation-maximization (EM) procedure \cite{tipping1999mixtures}. Interestingly, using the re-parameterization $\avect_k\leftarrow\sigma_k\avect_k$ and $\sigma^2_k\leftarrow 2\sigma^2_k$, it turns out that these EM updates match those of the blind source separation model proposed in \cite{sawada2007two}, up to a small additive constant for $\sigma^2_{k}$. A key difference is that in \cite{sawada2007two}, the observations are normalized so that $\|\xvect_t\|_2^2=1$. As such, \cite{sawada2007two} belongs to the class of spatial-feature clustering-based methods, similarly to DUET \cite{rickard2007duet}, while our method operates in the signal domain.

\vspace{-2mm}
\subsection{Parameter estimation via generalized moment matching}
\vspace{-2mm}
\label{sec:sketch}

In the general case $\alpha\neq 2$, estimation is done by generalized moment matching, that is, minimizing the difference between the empirical and theoretical values of a finite number of generalized moments, which are here samples of the empirical characteristic function of the data at some frequency vectors $\omegavect_j \in \mathbb{C}^M$, $j\in [1\dots J]$, to be matched with their analytical expression~\eqref{eq:xgivenw}. Following the methodology in \cite{sketching2016}, the vectors $\omegavect_j$ are drawn randomly according to some probability distribution $\Lambda$, in practice designed automatically from the data using the method prescribed in \cite{sketching2016}.

More precisely: given the data points to cluster $\xvect_1,\dots,\xvect_T \in \mathbb{C}^M$ (where the index $f$ has been dropped), the estimation is performed as follows:
\begin{enumerate}[leftmargin=*]
\item Draw $m$ random vectors $\omegavect_j \stackrel{iid.}{\sim} \Lambda$ for $j\in [1...J]$;
\item Compute the empirical characteristic function at these frequencies $\yvect = \left[\frac{1}{T}\sum_{t=1}^T e^{i\text{Re}(\omegavect_j^\star\xvect_t)}\right]_{j =1}^J \in \mathbb{C}^J$;
\item Estimate the model parameters \eqref{eq:modelparam} by (approximately) solving
\begin{equation}
\label{eq:momentmatching}
\min_{\thetavect} \left\lVert \yvect - \left[\psi_{\footnotesize\xvect|z=k}(\omegavect_j)\right]_{j=1}^J\right\rVert_2^2
\end{equation}
where $\psi_{\footnotesize\xvect|z=k}(\omegavect)$ is defined by \eqref{eq:xgivenw}, parameterized by $\thetavect$.
\end{enumerate}

\noindent\textbf{CL-OMPR.} The minimization \eqref{eq:momentmatching} is carried out by a modified version of the CL-OMPR algorithm \cite{sketching2016} adapted to our model. It is a greedy, heuristic algorithm precisely designed to perform mixture model estimation by generalized moment matching. Although it offers limited theoretical guarantees except for very particular settings \cite{Boyd2015}, it has been empirically shown to perform well for a large variety of models \cite{sketching2016}. In particular, it is applicable as soon as the considered mixture model has \emph{a closed-form characteristic function} with respect to the parameters of the model, which is the case for mixture of $\alpha$-stables distributions. Although it was initially designed to perform mixture model estimation on large databases, we use it here mostly because the probability density function of the proposed model \eqref{eq:xgivenw} does not enjoy an analytical expression. This forbids the use of classical methods such as EM. To our knowledge, there is no other algorithm capable of estimating mixtures of multivariate $\alpha$-stable distributions in the literature.

The CL-OMPR algorithm is a variant of Orthogonal Matching Pursuit (OMP), a classical greedy algorithm in compressive sensing. Like OMP, it iteratively \emph{adds} a component to the mixture model by maximizing its correlation to the residual signal. Since the space of parameters is continuous, this is done here with a gradient ascent randomly initialized. Furthermore, CL-OMPR alternates this greedy step with a non-convex, global gradient descent on \eqref{eq:momentmatching} initialized with the current support. This additional step adjust the whole support when a component is added. Finally, it also performs more iterations than OMP and includes a hard thresholding step to maintain the number of components at $K$, to allow for replacing spurious components. 

The CL-OMPR algorithm is described in detail in \cite{sketching2016}, where it is applied to Gaussian Mixture Model (GMM) estimation. Replacing the GMM by our $\alpha$-stable model is easily implemented and only requires computation of the gradient of $\psi(\xvect|z=k)$ with respect to the different parameters. The code is available at \url{https://github.com/nkeriven/alpha_stable_bss}.

\noindent\textbf{Approximate clustering.} A drawback of the $\alpha$-stable model to investigate in future work is that the pdf $p(\xvect|z=k)$ does not have an explicit expression. Therefore, the clustering of data points $\xvect_t$ cannot be done by exactly maximizing the posterior $p(\xvect_t|z=k)$ with respect to $k$. In other words, once we have estimated the mixture of $\alpha$-stable distributions, it is difficult to actually \emph{assign} each point to a component of the mixture.

Although a few methods may exist to approximately compute this posterior using approximate numerical integration \cite{Nolan2013}, in practice we found them to be extremely unstable and time consuming. Instead, we decided to cluster the data \emph{as if the model was Gaussian}, $i.e.$ with $\alpha_k=2$, since the likelihood is then computable.  Hence, the ``clustering'' part (and therefore the final source separation step) of both EM (Section \ref{sec:alpha2}) and the $\alpha$-stable model are in fact \emph{the same}. The difference between the two lies in the estimation of parameters $(\avect_k,\sigma^2_k,\pi_k)$. Our hope is that by using the more realistic $\alpha$-stable source model, steering vectors $\avect_k$ will be estimated more precisely.

\vspace{-2mm}
\subsection{Frequency permutation ambiguity}
\vspace{-2mm}
\label{sec:permut}
Once clustering is performed at each frequency, a permutation ambiguity remains as the assignment of frequency masks to sources is not known. This is a classical problem in blind source separation referred to as \textit{permutation alignment}. It notably occurs when using ICA \cite{hyvarinen2004independent} and clustering-based methods \cite{sawada2007two,duong_TSALP2010}. A number of techniques have been proposed to tackle it, based on temporal activation patterns \cite{sawada2007two}, steering vector models \cite{duong_TSALP2010} or adjacent frequency bands similarity \cite{di2016using}. The selection and tuning of a specific permutation technique highly depends on the type of signal and mixing model considered, which is out of the scope of this study. For this reason and for fairness, all methods evaluated in the next section benefited from the same \textit{oracle permutation} scheme. At each frequency, the permutation minimizing the mean-squared error between estimated and true source images is selected.

\vspace{-2mm}
\section{Evaluation and Results}
\vspace{-2mm}
We use two datasets for evaluation. First, a subset of the QUASI database\footnote{\scriptsize \url{www.tsi.telecom-paristech.fr/aao/en/2012/03/12/quasi/}} consisting in 10 musical excerpts of 30s. For each excerpt, we produced stereo ($M=2$) mixes of $K=4$ musical tracks (vocals, bass, drums, electric guitar, keyboard,...) using random pure gains and delays. Second, the TIMIT speech database\footnote{\scriptsize \url{catalog.ldc.upenn.edu/ldc93s1}}, from which we created 10 tracks of 30s. For each experiment we mix $K=3$ of them selected at random into $M=2$ channels, again with random pure gains and delays. In all cases, the gain difference between the two channels are at most 5dB and the delay is at most 20 samples. Note that none of the tested methods make assumption on the specific convolutive filters used for mixing, as long as they are relatively small compared to the Fourier analysis window. The STFT parameters were fixed to 64ms Hamming windows at 16kHz with $75\%$ overlap.

Each experiment is averaged over 100 trials: each of the 10 songs is selected 10 times, and at each speech trial random utterances are picked from TIMIT and mixed. The results are evaluated using the classical \texttt{bss\_eval} toolbox \cite{Vincentbsseval06}. They are expressed in terms of the signal-to-distortion ratio (SDR) and signal-to-interference ratio (SIR), evaluating the quality of the reconstructed source signals, and the mixing error ratio (MER), defined in \cite{vincent20092008}, evaluating the estimation of the steering vectors $\avect_k$.
%
%

\begin{table}
 \centering
 \subfloat[QUASI database (music), $K=4$]{
 \begin{tabular}{c|c|c|c|}
\cline{2-4}
& SDR (dB) & SIR (dB) & MER (dB) \\ \hline
\multicolumn{1}{|c|}{Mix} & $-5.96 \pm 4.96$ & $-5.49 \pm 4.85$ & N/A \\ \hline
\multicolumn{1}{|c|}{Oracle} & $8.33 \pm 3.16$ & $18.3 \pm 4.13$ & N/A \\ \hline\hline
\multicolumn{1}{|c|}{\cite{sawada2007two}} & $1.26 \pm 2.44$ & $2.88 \pm 3.82$ & $10.5 \pm 9.84$ \\ \hline
\multicolumn{1}{|c|}{EM} & $3.50 \pm 2.87$ & $9.04 \pm 4.92$ & $12.3 \pm 11.0$ \\ \hline
\multicolumn{1}{|c|}{CF-GMM} & $3.80 \pm 2.53$ & $8.60 \pm 3.62$ & $12.3 \pm 9.90$\\ \hline
\multicolumn{1}{|c|}{CF-$\alpha$} & $\mathbf{4.11 \pm 2.59}$ & $\mathbf{9.17 \pm 3.51}$ & $\mathbf{12.7 \pm 9.73}$ \\ \hline
\end{tabular}
\label{fig:subquasi}
}
\\
 \subfloat[TIMIT database (speech), $K=3$]{
 \begin{tabular}{c|c|c|c|}
\cline{2-4}
& SDR (dB) & SIR (dB) & MER (dB) \\ \hline
\multicolumn{1}{|c|}{Mix} & $-3.14 \pm 1.91$ & $-3.13 \pm 1.90$ & N/A \\ \hline
\multicolumn{1}{|c|}{Oracle} & $11.9 \pm 0.98$ & $25.9 \pm 1.05$ & N/A \\ \hline\hline
\multicolumn{1}{|c|}{\cite{sawada2007two}} & $2.16 \pm 1.33$ & $4.90 \pm 2.54$ & $\mathbf{22.0 \pm 6.57}$ \\ \hline
\multicolumn{1}{|c|}{EM} & $0.54 \pm 0.50$ & $1.44 \pm 1.21$ & $12.0 \pm 3.64$ \\ \hline
\multicolumn{1}{|c|}{CF-GMM} & $1.60 \pm 1.10$ & $4.13 \pm 2.46$ & $14.8 \pm 3.32$\\ \hline
\multicolumn{1}{|c|}{CF-$\alpha$} & $\mathbf{2.70 \pm 1.74}$ & $\mathbf{6.11 \pm 3.31}$ & $18.9 \pm 2.72$ \\ \hline
\end{tabular}
\label{fig:subtimit}
}
\caption{Separation results with $K$ sources and $M=2$ channels, for the four clustering algorithms as well as oracle and mixture results. Each slot contains the mean and standard deviation over the $100$ trials and $K$ sources, $i.e.$ over $100K$ values.\vspace{-5mm}}
\label{tab:sep}
\end{table}

We compare the following 4 clustering algorithms (recall that in each case, binary masks are created using the oracle permutation method of Sec. \ref{sec:permut}):
\vspace{-2mm}
\begin{itemize}[leftmargin=*]
\item {\bf EM}: The clustering is done with a GMM as described in Sec. \ref{sec:alpha2}. The EM algorithm is repeated $10$ times and parameters yielding the best log-likelihood are kept.
\item {\bf \cite{sawada2007two}}: This is our implementation of the method of Sawada et al. using normalized observation, as described in Section \ref{sec:alpha2}. The EM is also repeated 10 times.
\item {\bf CF-GMM}: the clustering is formed with the moment matching method of Sec. \ref{sec:sketch}, but with all the $\alpha_k$ fixed to $2$. Hence, both EM and CF-GMM achieve estimation in a Gaussian setting, but with different cost functions: while EM maximizes likelihood, CF-GMM performs generalized moment matching of the characteristic function (CF).
\item {\bf CF-$\alpha$}: the clustering is done with the mixture of $\alpha$-stable distributions of Sec. \ref{sec:sketch}. As mentioned before, recall that the \emph{clustering} part is done by approximating the model as Gaussian, only the \emph{estimation} of the parameters is different.
\end{itemize}
\vspace{-2mm}
To put the results in context, we also outline the ``best'' and ``worst'' possible results. In {\bf oracle}, the separation is performed with the binary mask formed by considering the source that has the highest energy at each TF bin (with oracle knowledge of each source signal). In {\bf mix}, the result are obtained by directly feeding the mixture signal into the function \texttt{bss\_eval\_images}.

\noindent{\bf Separation results.} In Table \ref{tab:sep} we show the separation results for all algorithms. Recall that \cite{sawada2007two} performs separation purely based on spatial clustering while EM, CF-GMM and CF-$\alpha$ also rely on statistical source models. The results suggest that first, using source models is more beneficial in heavily underdetermined scenarios, \textit{e.g.} Table \ref{tab:sep}(a), where source signals are less sparse and more numerous. Second, the proposed $\alpha$-stable model is better suited than Gaussian models for both speech and music sources. Finally, the proposed approach blindly estimates mixing filters in a more stable way than the EM approach of \cite{sawada2007two} despite its multiple initialization, as showed by the lower standard deviations of MER.

\begin{figure}
\centering
\includegraphics[width =0.48\textwidth]{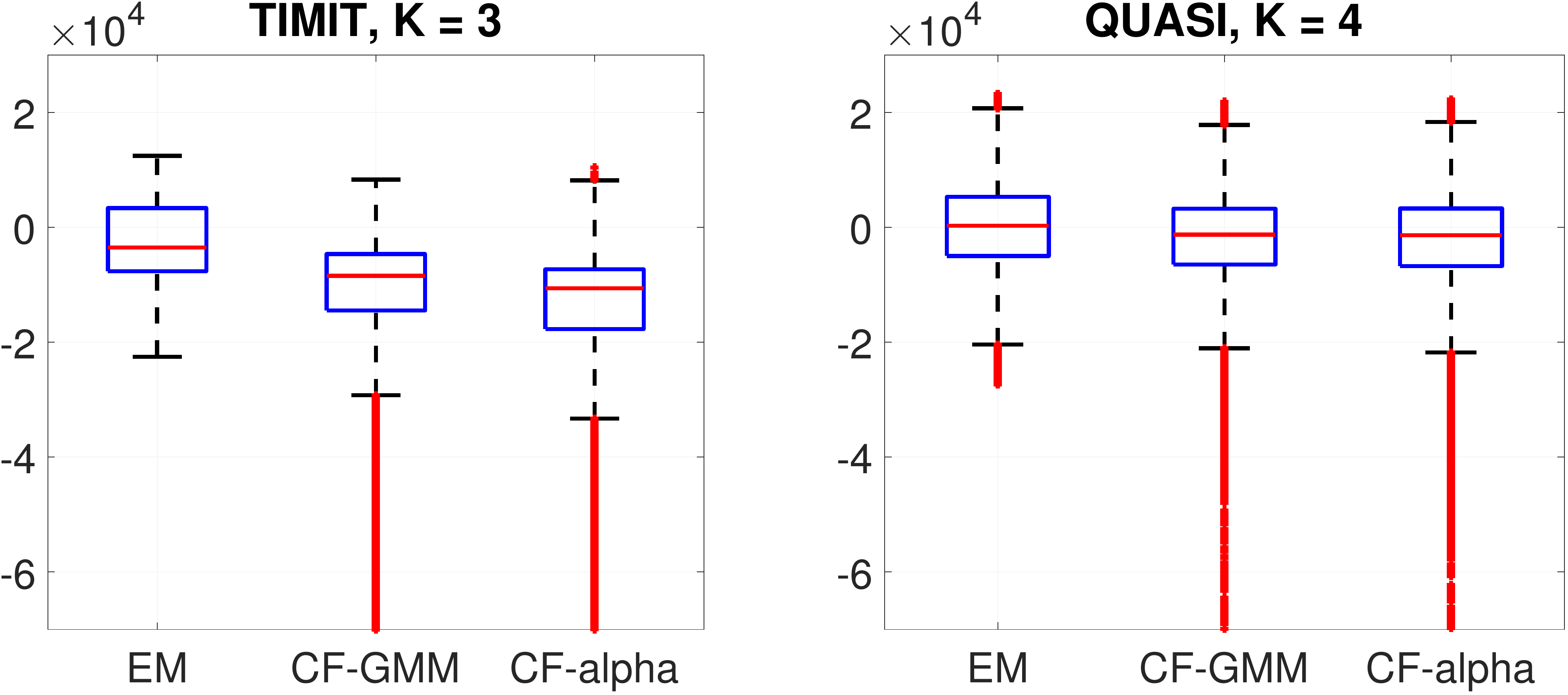}
\caption{Log-likelihood of the data at each frequency index for each trial (i.e. $100F$ values), for the EM, CF-GMM and CF-$\alpha$. For the latter, the ``likelihood'' is computed with $\alpha=2$ (Gaussian), even if a different $\alpha$ was estimated. For readability the low end of the $y$-axis has been cut at $-7.10^4$, the CF-GMM and CF-$\alpha$ algorithms have outliers that go down to, respectively, $-2.10^{10}$ and $-3.10^{10}$.\vspace{-2mm}}
\label{fig:likelihood}
\end{figure}

\noindent{\bf Relevance of log-likelihood.} A somewhat surprising observation is that CF-GMM significantly outperforms EM on speech data, despite the fact that both estimate a GMM.
In Fig. \ref{fig:likelihood} we compare the log-likelihood results obtained with the three algorithms during the clustering phase subsequent to the estimation of the parameters (recall that all three algorithms have the same \emph{clustering} phase that do not use the estimated $\alpha_k$). As expected, EM significantly outperforms the two other algorithms on this criterion. This is not surprising since EM aims at maximizing the log-likelihood while the two CF algorithms consider only the characteristic function. Since the CF approaches outperform EM in terms of separation results, we conclude that maximization of the log-likelihood, while natural, might not be the most appropriate approach to estimate the mixture parameters in this case, which is an interesting lead for future research. 

\vspace{-3mm}
\section{Conclusion}
\vspace{-2mm}
We presented a novel method for multichannel blind separation of audio sources using an $\alpha$-stable model for source signals, combined with the assumption that only one source dominates each $(t,f)$ point. The parameters of the proposed model, including distinct scale and $\alpha$ values for each source, are estimated at each frequency using a novel method based on random generalized moment matching. Results show that using oracle permutations, the proposed model performs better than Gaussian models, and that the proposed estimation method outperforms EM even using the same Gaussian model. Future work will further investigate the $\alpha$ and scale values estimated by our method. In particular, it would be interesting to see if they can be constrained or exploited to resolve permutation ambiguities. The potential of random generalized moment matching versus maximum likelihood methods in source separation should also be further studied.

\bibliographystyle{IEEEbib}
\ninept
\bibliography{refs_ICASSP2018_alphabss_rev}

\begin{thebibliography}{10}

\bibitem{ss_review2010}
P.~Comon and C.~Jutten, Eds.,
\newblock {\em Handbook of Blind Source Separation: Independent Component
  Analysis and Blind Deconvolution},
\newblock Academic Press, 2010.

\bibitem{Yilmaz2004}
O.~Yilmaz and S.~Rickard,
\newblock ``{Blind separation of speech mixtures via time-frequency masking},''
\newblock {\em IEEE Transactions on Signal Processing}, vol. 52, no. 7, pp.
  1830--1847, July 2004.

\bibitem{rickard2007duet}
Scott Rickard,
\newblock ``The duet blind source separation algorithm,''
\newblock {\em Blind Speech Separation}, pp. 217--237, 2007.

\bibitem{vincentSSoverview2014}
E.~Vincent, N.~Bertin, R.~Gribonval, and F.~Bimbot,
\newblock ``From blind to guided audio source separation: How models and side
  information can improve the separation of sound,''
\newblock {\em IEEE Signal Processing Magazine}, vol. 31, no. 3, pp. 107--115,
  May 2014.

\bibitem{hyvarinen2004independent}
A.~Hyv{\"a}rinen, J.~Karhunen, and E.~Oja,
\newblock {\em Independent component analysis}, vol.~46,
\newblock John Wiley \& Sons, 2004.

\bibitem{belouchrani1997blind}
A.~Belouchrani, K.~Abed-Meraim, J-F Cardoso, and E.~Moulines,
\newblock ``A blind source separation technique using second-order
  statistics,''
\newblock {\em IEEE Transactions on signal processing}, vol. 45, no. 2, pp.
  434--444, 1997.

\bibitem{benaroya-separation06}
L.~Benaroya, F.~Bimbot, and R.~Gribonval,
\newblock ``Audio source separation with a single sensor,''
\newblock {\em IEEE Transactions on Audio, Speech and Language Processing},
  vol. 14, no. 1, pp. 191--199, Jan. 2006.

\bibitem{duong_TSALP2010}
N.Q.K. Duong, E.~Vincent, and R.~Gribonval,
\newblock ``Under-determined reverberant audio source separation using a
  full-rank spatial covariance model,''
\newblock {\em IEEE Transactions on Audio, Speech and Language Processing},
  vol. 18, no. 7, pp. 1830 --1840, sept. 2010.

\bibitem{EM-superimposedFeder}
M.~Feder and E.~Weinstein,
\newblock ``Parameter estimation of superimposed signals using the {EM}
  algorithm,''
\newblock {\em IEEE Transactions on Acoustics}, vol. 36, pp. 477--489, 1988.

\bibitem{ozerov_generalframework2011}
A.~Ozerov, E.~Vincent, and F.~Bimbot,
\newblock ``A general flexible framework for the handling of prior information
  in audio source separation,''
\newblock {\em IEEE Transactions on Audio, Speech and Language Processing},
  vol. PP, no. 99, pp. 1, 2011.

\bibitem{nugraha2016multichannel}
A.~Nugraha, A.~Liutkus, and E.~Vincent,
\newblock ``Multichannel audio source separation with deep neural networks,''
\newblock {\em IEEE Transactions on Audio, Speech, and Language Processing},
  vol. 24, no. 9, pp. 1652--1664, 2016.

\bibitem{yoshii2016student}
Kazuyoshi Yoshii, Katsutoshi Itoyama, and Masataka Goto,
\newblock ``Student's t nonnegative matrix factorization and positive
  semidefinite tensor factorization for single-channel audio source
  separation,''
\newblock in {\em IEEE International Conference on Acoustics, Speech and Signal
  Processing (ICASSP)}, Shanghai, China, April 2016.

\bibitem{kidmose2001blind}
P.~Kidmose,
\newblock {\em Blind separation of heavy tail signals},
\newblock Ph.D. thesis, Technical University of Denmark, Lyngby, Denmark, 2001.

\bibitem{kidmose2001independent}
P.~Kidmose,
\newblock ``Independent component analysis using the spectral measure for
  alpha-stable distributions,''
\newblock in {\em IEEE-EURASIP 2001 Workshop on Nonlinear Signal and Image
  Processing}, 2001, vol. 400.

\bibitem{samoradnitsky1994stable}
G.~Samoradnitsky and M.~Taqqu,
\newblock {\em Stable non-Gaussian random processes: stochastic models with
  infinite variance}, vol.~1,
\newblock CRC Press, 1994.

\bibitem{nikias1995signal}
C.~Nikias and M.~Shao,
\newblock {\em Signal processing with alpha-stable distributions and
  applications},
\newblock Wiley-Interscience, 1995.

\bibitem{harmonizable2015}
A.~Liutkus and R.~Badeau,
\newblock ``Generalized {W}iener filtering with fractional power
  spectrograms,''
\newblock in {\em IEEE International Conference on Acoustics, Speech and Signal
  Processing (ICASSP)}, Brisbane, Australia, April 2015.

\bibitem{fontaine2017explaining}
M.~Fontaine, A.~Liutkus, L.~Girin, and R.~Badeau,
\newblock ``Explaining the parameterized wiener filter with alpha-stable
  processes,''
\newblock in {\em IEEE Workshop on Applications of Signal Processing to Audio
  and Acoustics (WASPAA)}, 2017.

\bibitem{csimcsekli2015alpha}
U.~{\c{S}}im{\c{s}}ekli, A.~Liutkus, and A.T. Cemgil,
\newblock ``Alpha-stable matrix factorization,''
\newblock {\em IEEE Signal Processing Letters}, vol. 22, no. 12, pp.
  2289--2293, 2015.

\bibitem{liutkusFIM2015}
A.~Liutkus, T.~Olubanjo, E.~Moore, and M.~Ghovanloo,
\newblock ``{Source Separation for Target Enhancement of Food Intake Acoustics
  from Noisy Recordings},''
\newblock in {\em {IEEE Workshop on Applications of Signal Processing to Audio
  and Acoustics (WASPAA)}}, New Paltz, NY, United States, Oct. 2015.

\bibitem{sketching2016}
N.~Keriven, A.~Bourrier, R.~Gribonval, and P.~P{\'{e}}rez,
\newblock ``Sketching for large-scale learning of mixture models,''
\newblock {\em arXiv:1606.02838, Information and Inference: A Journal of the
  IMA}, 2017.

\bibitem{gribonval2017compressive}
R.~Gribonval, G.~Blanchard, N.~Keriven, and Y.~Traonmilin,
\newblock ``Compressive statistical learning with random feature moments,''
\newblock {\em arXiv preprint arXiv:1706.07180}, 2017.

\bibitem{fontaine2017scalable}
M.~Fontaine, C.~Vanwynsberghe, A.~Liutkus, and R.~Badeau,
\newblock ``Scalable source localization with multichannel alpha-stable
  distributions,''
\newblock in {\em 25th European Signal Processing Conference (EUSIPCO 2017)},
  2017.

\bibitem{circularSymmetricGallager}
R.~Gallager,
\newblock ``{Circularly Symmetric Complex Gaussian Random Vectors - A
  Tutorial},''
\newblock Tech. {R}ep., Massachusetts Institute of Technology, 2008.

\bibitem{sawada2007two}
Hiroshi Sawada, Shoko Araki, and Shoji Makino,
\newblock ``A two-stage frequency-domain blind source separation method for
  underdetermined convolutive mixtures,''
\newblock in {\em Applications of Signal Processing to Audio and Acoustics,
  2007 IEEE Workshop on}. IEEE, 2007, pp. 139--142.

\bibitem{tipping1999mixtures}
Michael~E Tipping and Christopher~M Bishop,
\newblock ``Mixtures of probabilistic principal component analyzers,''
\newblock {\em Neural computation}, vol. 11, no. 2, pp. 443--482, 1999.

\bibitem{Boyd2015}
Nicholas Boyd, Geoffrey Schiebinger, and Benjamin Recht,
\newblock ``{The Alternating Descent Conditional Gradient Method for Sparse
  Inverse Problems},''
\newblock pp. 1--21, 2015.

\bibitem{Nolan2013}
John~P. Nolan,
\newblock ``{Multivariate elliptically contoured stable distributions: Theory
  and estimation},''
\newblock {\em Computational Statistics}, vol. 28, no. 5, pp. 2067--2089, 2013.

\bibitem{di2016using}
Leandro~E Di~Persia and Diego~H Milone,
\newblock ``Using multiple frequency bins for stabilization of fd-ica
  algorithms,''
\newblock {\em Signal Processing}, vol. 119, pp. 162--168, 2016.

\bibitem{Vincentbsseval06}
E.~Vincent, R.~Gribonval, and C.~F\'evotte,
\newblock ``Performance measurement in blind audio source separation,''
\newblock {\em IEEE Transactions on Audio, Speech and Language Processing},
  vol. 14, no. 4, pp. 1462 --1469, July 2006.

\bibitem{vincent20092008}
Emmanuel Vincent, Shoko Araki, and Pau Bofill,
\newblock ``The 2008 signal separation evaluation campaign: A community-based
  approach to large-scale evaluation,''
\newblock in {\em International Conference on Independent Component Analysis
  and Signal Separation}. Springer, 2009, pp. 734--741.

\end{thebibliography}

\end{document}